\documentclass[runningheads]{llncs}
\pdfoutput=1
\usepackage{graphicx}
\usepackage{algorithm}
\usepackage{algorithmic}
\usepackage{geometry}

\usepackage{makeidx}  

\graphicspath{{Images/}}

\begin{document}

\title{Cursive Scene Text Analysis by Deep Convolutional Linear Pyramids}
%
%
\author{Saad Bin Ahmed\inst{1,2} \and Saeeda Naz\inst{4} \and
Muhammad Imran Razzak\inst{3}\thanks{Corresponding author. imran.razzak@ieee.org}
 \and Rubiyah Yusof\inst{2}}
\authorrunning{S. B. Ahmed et al, (Pre-print version, Presented in ICONIP-2018)}
%
\institute{King Saud bin Abdulaziz University for Health Sciences, Riyadh, Saudi Arabia \\ \email {\{ahmedsa\}@ksau-hs.edu.sa} \\ \and
Malaysia Japan International Institute of Technology (MJIIT), Universiti Teknologi Malaysia, Kuala-Lumpur, Malaysia \\ \email{\{saad2,rubiyah.kl\}@utm.my} \\  \and
University of Technology, Sydney, Australia  \\\email{\{imran.razzak@ieee.org\} } \\ \and
Govt. Post graduate College No. 01, Higher Education Department, Abbottabad, Pakistan \\\email{\{saeedanaz292\}@gmail.com}}

\maketitle 
\begin{abstract}
The camera captured images have various aspects to investigate.
Generally, the emphasis of research depends on the interesting regions. 
Sometimes the focus could be on color segmentation, object detection or scene text analysis.
The image analysis, visibility and layout analysis are the tasks easier for humans as suggested by behavioural trait of humans, but in contrast when these same tasks are supposed to perform by machines then it seems to be challenging.
The learning machines always learn from the properties associated to provided samples.
The numerous approaches are designed in recent years for scene text extraction and recognition and the efforts are underway to improve the accuracy.  
The convolutional approach provided reasonable results on non-cursive text analysis appeared in natural images.
The work presented in this manuscript exploited the strength of linear pyramids by considering each pyramid as a feature of the provided sample. 
Each pyramid image process through various empirically selected kernels. 
The performance was investigated by considering Arabic text on each image pyramid of EASTR-42k dataset.
The error rate of $0.17$\%  was reported on Arabic scene text recognition.
\end{abstract}

\keywords{Linear Pyramids, Kernels, Feature Extraction, Arabic Scene Text.}

\section{Introduction}
The text localization is perceived as a fundamental step in scene text recognition.
Although few approaches for text localization have been suggested by pattern recognition and computer vision research communities, more effort is still required to define text localization approaches for cursive scene text.
The localization techniques are not usually specific to a particular script.
One successful approach can be applied on number of scripts having the same nature image.
In presented work, the analysis is performed on Arabic cursive script.
Later, applied localization technique which has proved encouraging results on EASTR-42k dataset.

In scene text image analysis, features play a prominent role in correct recognition of given scene text. 
The heterogeneous features increased prospect of obtaining better accuracy. 
The scene text image contains non-hierarchical features that are extracted from local regions.
These features are described as distinctive properties which need to quantize in standard format by using a spatial or statistical techniques. 
The text represented in scene images contain inter and intra-class variations, this difference is comparatively easier to handle in Latin rather than in Arabic.
The characteristics of Arabic and Arabic-like scripts are writing direction (i.e., from right to left), representation of some characters with diacritical marks and appearance of same character with respect to its position makes this script more challenging.
Although, recent research on printed and handwritten Arabic-like scripts presented state-of-the-art results as reported by~\cite{urdu1,urdu2,asar2,cnn-urdu} \cite{ahmed2017ucom,naz2016urdu,naz2016segmentation,razzak2010hmm} but scene text recognition of Arabic script is still in its early stages \cite{asar2}.
The scene text recognition techniques fall into texture, component and hybrid based methods~\cite{asar2}. 
This paper presents texture based methods which help in localizing the text based on filtration techniques.

The background of image is discriminated by the color and lighting effects. 
The discrimination usually provides a clue about the presence of a text in a particular region. 
This fact is exploited and established the steps about how this dicriminative property can be helpful in the determination of text localization. 
This paper proposed a feature extraction technique that helps in Arabic scene text recognition using linear spatial pyramids based on image analysis filters. 
The Gaussian pyramid is applied for smoothing the images.
Every subsequent image passes through various image processing filters. 
The convolutional approach is applied on image pyramids which helps in developing feature vectors of each image. 
Every pyramid image is taken into account and passed through the filtration and convolution processes.

The extracted feature vectors were given to Multi-Dimensional Long Short Term Memory (MDLSTM) networks for a purpose to learn the vector sequence. 
The MDLSTM is an appropriate approach for sequence learning tasks and have been successfully applied on numerous sequence learning cursive text recognition during recent years~\cite{urdu1,urdu2,cnn-urdu}.
The contribution of proposed work is alienated as follows,
\begin{enumerate}
    \item The linear image pyramid technique has been adapted and applied on localization of Arabic scene text.
    \item A novel way is proposed to determine the text location and extract text from given scene image. To accomplish this, the presented work applies image processing filters like laplacian, large blur, small blur, sobel\_x, sobel\_y and sharping the image text and convolving all of them with empirically selected kernel. 
    \item Deep MDLSTM architecture is employed to see the performance of presented method which demonstrated very encouraging results.
\end{enumerate}

Section $2$ describe the proposed methodology which further explain the formulation of linear pyramids by image filters. 
The experimental analysis is elaborated in section $3$, whereas, conclusion is summarized in section $4$.  

\section{Related Work}
The research on scene text analysis has increased to manifold for since few years.
There are numerous techniques proposed for text localization, extraction and recognition from natural images.
Text analysis in scene text is not an easy task because it involves a lot of other factors in an image that may create complication in smooth execution of specifically designed techniques. 
In this section some latest techniques have been compiled that suggest solutions for text localization and extraction from scene image. 

Scene text does not have only one type of feature associated with it, instead it has numerous features that may play a conclusive role in correct determination of character/ words.
Therefore, there is a need to investigate more about the features that may be distinct in nature but helpful in learning and recognition process.
One such work is presented by Lazebnik etal~\cite{py1}.
They proposed a holistic approach for image categorization that uses pyramid matching kernels to demonstrate the performance of their proposed architecture.
They evaluated their proposed technique on fifteen categories using Graz dataset~\cite{py1}.
The Graz dataset contains highly discriminative images and their proposed method provides good accuracy on image categorization. 
Another dataset they evaluated on Caltech-101 wasthe one which contains $31-800$ images per category. 
This dataset is considered as most diverse dataset provided for research purpose.
The experimental setting was depicted in their manuscripts~\cite{py2,py3}.
They trained $30$ images per class and the test set contains $50$ images per class.
They reported good accuracy by using their dataset as can seen further detail in their paper.

Another presented work is about spatial pyramid matching based on invariant features using sparse coding method presented by\cite{py4}, they are using max pooling strategy in the histogram on multiple spatial scales to subsume translation and scale invariance. 
Their work is an extension of the the work presented in \cite{py1}.
Furthermore, they presented the results obtained on Caltech-256, 5 images and TRECVID 2008 surveillance video images.
They also compared their results by evaluating their technique on the same datasets as reported in \cite{py1}.
They implement three variation of spatial pyramid matching (SPM) algorithm. 
The first variation is SPM with Chi square kernels.
Another variation is to use linear kernel on spatial pyramid histogram.
The last variation, by using it they reported best accuracy is inclusion of Scale Invariant Feature Transformation (SIFT) features with combination of linear kernel on spatial pyramids.
The detail about performed experiments can be found in their respective manuscript.

The spatial pyramid model presents very interesting results by incorporating other techniques for a purpose to enhance the performance.
One such technique is presented by~\cite{py5}, they proposed two different models with incorporation of spatial pyramid knowledge.
Their first model is an image independent model while the second one adapted the image contents.
They used PASCAL VOC 2007, 2008, and 2009 dataset for evaluation of their proposed models. 
Their average results vary from $62.5$\% - $66.5$\% as reported.

Although application of spatial linear pyramid is new for scene text recognition problems. 
SPM has not been applied before for categorization of text related problems. 
This paper is presenting a novel solution for text localization problem.
As this technique has been comparatively applied successfully, on number of object recognition in scene images as can witnessed in~\cite{py1,py2,py5,py6}.
The presented work is exploiting its potential in the field of text localization.

\section{Proposed Methodology}
The linear pyramids proved very convincing results on image categorization in natural images.
The strength of under discussion technique has yet to explore regarding text categorization in natural images.
This idea is a focal point of presented research that will try to explore how this technique can be beneficial in determining correct text localization.

\begin{figure}[!htb]
\centering
\includegraphics[scale=0.5]{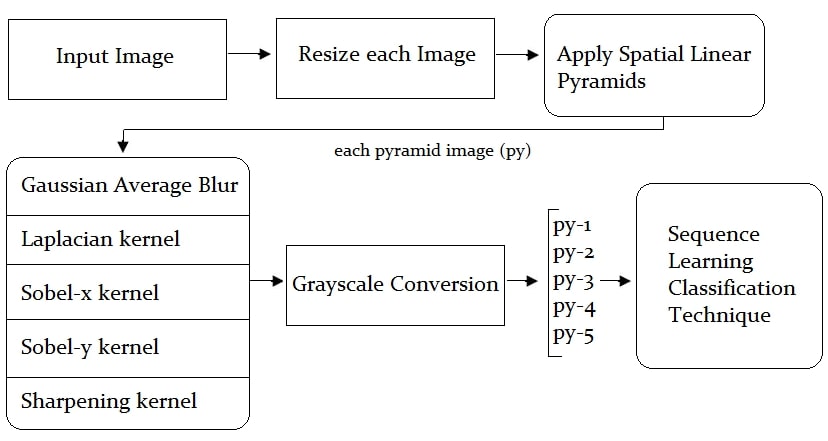}
\caption{Flowchart of proposed architecture}
\label{neuro-method}
\end{figure}

\begin{figure}
\centering
\includegraphics[scale=0.5]{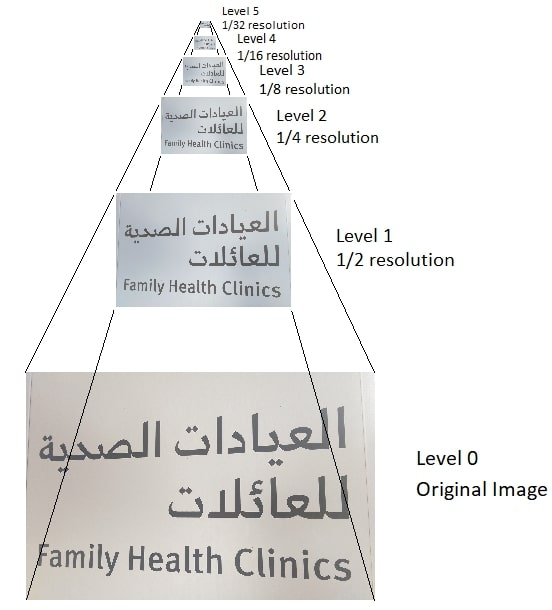}
\caption{ Visual representation of Linear spatial pyramids with 6 levels.}
\label{neuro-1}
\end{figure}

This section provides description about the implementation of proposed idea.
Figure~\ref{neuro-method} elaborates proposed idea in detail. 
Every image is rescaled into standard size.
After that linear pyramid's method is applied.
The linear pyramid is described in $6$ levels as indicated in Figure~\ref{neuro-1}.
The linear pyramid generates $5$ images of different resolution.
Each image passes through filter pack and the resultant images are converted into grayscale. 
At the end, all pyramid gray scale images were given to classifier for training purpose.
The whole process defined into following subsections.

\subsection{Formulation of linear spatial pyramids of Cursive Arabic scene text}

The image pyramid contemplated as a two dimensional arrays which meant to represent image from smaller to smallest size reducing the image information at each level from base to the top of pyramid.
There are numerous ways suggested to define pyramids as explained in~\cite{agent}, but in-practice, procedure to define pyramids is considered a step from base to it's top.

In this work, RGB values are accounted where each cell contains the percentage of $R$, $G$, and $B$ colors.
The reason for proposing pyramids for text localization is its ability to group the text image/ words in an appropriate resolution so it may contribute in depicting feature as a whole in absence of complex computation. 
In today's era, systems are more intelligent in processing the acquired image, but difficult to understand the content represented in an image.
The human perception of recognizing text presented in various fonts regardless of their sizes inspired the machine models where its realization is implemented by linear pyramids.

The input image is passed through the loop where Gaussian pyramid is applied on each image.
Each image is resized to a specific height which should not be greater than $30$. 
The height limit described as the minimum size of the image considered in conducted experiments.

\subsection{Preprocessing of image pyramids by Image filters}
At first, the irrelevant and unnecessary information is removed from the image so that it should not merge with the textual objects during pyramid building. 
The text should be in focus during the process that's why there is a need to eliminate such patterns that may confuse the recognition process.
\begin{figure}[!htb]
\centering
\includegraphics[scale=0.3]{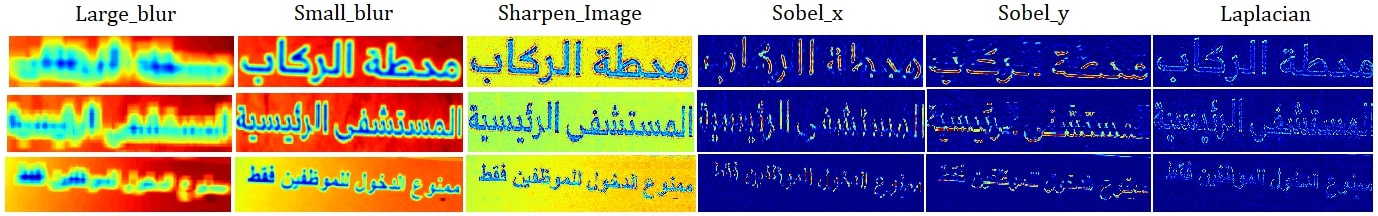}
\caption{Arabic scene text pre-processing with good examples.}
\label{neuro-good}
\end{figure}

\begin{figure}[!htb]
\centering
\includegraphics[scale=0.3]{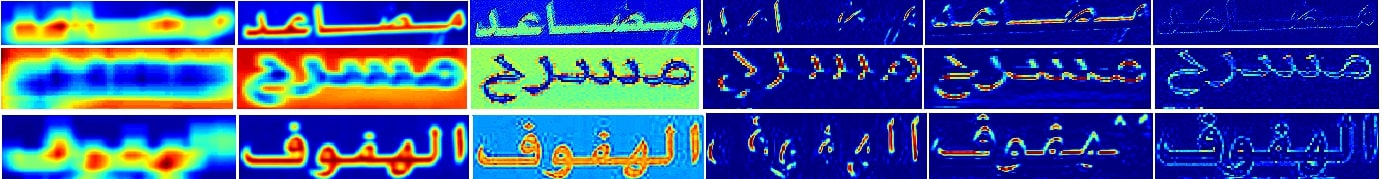}
\caption{Arabic scene text examples where segmentation is misclassified}
\label{neuro-mis}
\end{figure}

The previous section mentioned about how image pyramid was constructed, but the description about how it may be beneficial for recognition process not specified. 
This section explicitly concentrate on how the pyramid images may be used for feature extraction.
Figure \ref{neuro-good} shows good examples where text has categorized correctly after passing it through number of filters that contribute in feature building. 
Figure \ref{neuro-mis} represents misclassified text which degrades the recognition accuracy.

\subsection{LSTM and it's implicit segmentation}
The cursive text is more complex in nature in comparison to Latin. 
In such text, sequence is very important to know. 
In the cursive scripts like Arabic where each character has four representation without sharing any similarities among the shapes, the sequence learning is an appropriate choice for learning the temporal sequential behavior. 
In the scripts where segmentation is tedious to perform, an implicit segmentation is a suitable approach for such problems.
In presented work of Arabic or Arabic-like script analysis to date, most authors proposed their methodology by using implicit segmentation which is embedded in their presented technique.
One such technique is called Long Short Term Memory (LSTM) network which is a recurrent neural network approach, designed for sequence learning with implicit segmentation.

By implicit approach, the segmentation and recognition of characters are accomplished at the same time. 
There is a trade-off  between the numbers of segments of a given word. 
The computation time also plays an important role with respect to number of segments. 
Larger the number of segments of a given word, the more time it requires for computation. 
In that particular situation, the hypothesis that is made for recognition of a given character increases but at the expense of compromised computation time. 
On the other hand, less number of segments may produce result in ideal time, but in that specific situation, recognition results may suffer. 

\begin{figure}[!htb]
\centering
\includegraphics[scale=0.7]{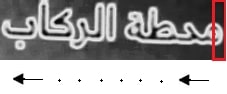}
\caption{ Gray scale image of x-height 60 x 1, moving from right to left.}
\label{neuro-5}
\end{figure}

Regarding the problems where sequence is important, there is a need to require powerful method that may learn context and make the task easier for recognition. 
Due to built-in complexities in Arabic scripts, it is usually impossible to segment cursive characters by explicit means.
The tentative segmentation of word may predict over segmentation and loop determination problem~\cite{alex}.
In over segmentation, the image may heuristically be over segmented with the horizontal distance in width.
The input image is adjusted by x-height which is $60$ pixels as shown in Figure~\ref{neuro-5}.

In Arabic or Arabic-like scripts, as mentioned earlier that characters may change its shape according to its position in a word. 
The traditional segmentation techniques can not apply on Arabic scripts \cite{naz2016arabic,naz2014optical}, for this purpose there is a need to look on implicit segmentation that may follows probabilistic or statistical measure for segmentation of such complex cursive scripts. 

\section{Experimental Analysis}
This section is presenting the experimentation analysis performed by implementing the proposed architecture. 
Experimental study was carried out by consideration of following points,
\begin{enumerate}
    \item Each pyramid image participates individually to inquire about which image size of the proposed architecture presented good results.
    \item The number of hidden layers contribute in chalking out of the best accuracies.
   
\end{enumerate}

Following sub section explains about proposed dataset used for experimentation. 

\subsection{Scene Text Dataset}
The dataset has a great importance for evaluating the proposed techniques or methods.
The dataset is prepared to demonstrate the performance of presented technique.
The dataset for Arabic camera captured images have not been developed yet.
The prime concern is to have scene images having Arabic text in focus, segment them and use them to validate the research tasks.
The bilingual scripts is common in most of gulf countries which prompts us to present dataset for English as well in addition to Arabic. 
In dataset, the taken samples tried to cover maximum variability of Arabic scene text data.
There are various text font sizes, styles and colors appeared in constraint environment which have been captured for research purpose.
The proposed dataset is disintegrated into Arabic and English scene text samples, as indicated in Table~\ref{table:1}.

\begin{table}
\centering
 \caption{EASTR-42K division based on complexity.}
 \label{table:1}
\begin{tabular}{ |c|c|c|c| }

 \hline
 \multicolumn{4}{|c|}{EASTR-42K Dataset} \\
 \hline
  \textbf{Language}&    \textbf{textlines}& \textbf{words}&\textbf{characters}\\ \hline
     
Arabic & 8,915 & 10,593 & 16,000\\
 \hline
English & 2,601  & 5,172 & 7,390 \\
  \hline
\end{tabular}
\end{table}

The prime concern of this manuscript is to investigate the performance of Arabic scene text proposed architecture.

\begin{table}[!t]
\caption{Experimentation Results reported by considering each Image Pyramid.} \label{tab:py}
\setlength\tabcolsep{0pt} 
\smallskip
\centering
\begin{tabular}{ |c|c|}
\hline

\textbf{Pyramid Image} & \textbf{Arabic}  \\
\hline
py-1  & 77.30 $\pm$ 0.45   \\ \hline
py-2  & 73.55 $\pm$ 0.63 \\\hline
py-3  & 68.21 $\pm$ 2.17 \\\hline
py-4  & 56.73 $\pm$ 0.59  \\\hline
py-5  & 49.71 $\pm$ 0.81  \\
\hline

\end{tabular}
\end{table}

The best accuracy is computed on py-1.
The established hypothesis is that large images have discrete information which posses image features that are used to improve the accuracy.

\begin{table}[!htb]
\caption{F-measure score observed on Arabic samples.} \label{tab:fa}
\setlength\tabcolsep{0pt} 
\smallskip
\centering
\begin{tabular}{ |c|c|c|c| }
\hline

\textbf{~Pyramid Images~} & \textbf{~Precision~} & \textbf{~Recall~} & \textbf{~F-measure~} \\
\hline
py-1  & 0.82  & 0.74 & 0.78  \\ 
py-2  & 0.78 & 0.70 & 0.74\\
py-3  & 0.69 & 0.58 & 0.63\\
py-4  & 0.60 & 0.47 & 0.53 \\
py-5  & 0.58 & 0.45 & 0.51 \\
\hline
\end{tabular}
\end{table}

The experimentation analysis is computed by f-measure on Arabic samples as summarized in Table \ref{tab:fa}.
It has been observed the accuracy gets lower and lower when pyramid gets small. 
The hypothesis is that large image contains detailed information which contributes to accuracy.

\begin{figure}[!htb]
\centering
\includegraphics[scale=0.4]{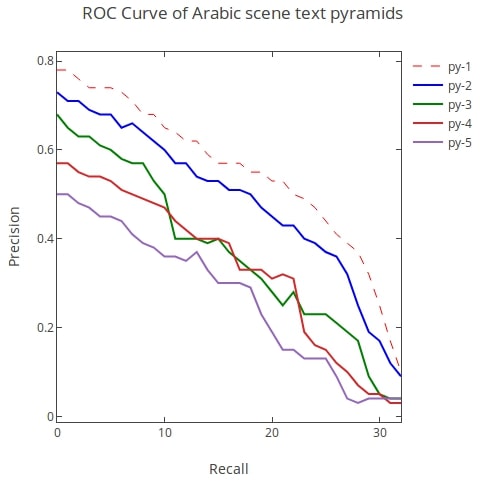}
\caption{ Precision/ Recall curve on Arabic samples.}
\label{neuro-9}
\end{figure}

Table~\ref{tab:whole} depicts the recognition results reported on Arabic by considering all pyramids of single image as a feature vector.
As observed the recognition is improved because every time network initial value is guesstimated which may improve the network learning as observed in the table below.

\begin{table}[!t]
\caption{Experimentation results reported by considering whole image pyramids.} \label{tab:whole}
\centering
\begin{tabular}{ |c|c| }
\hline
\textbf{Scripts} & \textbf{Text pyramids}  \\
\hline
Arabic  & 82.71 $\pm$ 1.13  \\ 
\hline

\end{tabular}
\end{table}

Table~\ref{tab:exp}, presents the experimental observation calculated on different hidden layer units. 
As observed, the best accuracy was reported when there are $100$ hidden layer units.
There is an observation that if reasonable memory units were provided during network learning, it can handle the complex input appropriately. 
The reasonable memory unit selection is empirically selected as explained in performed experiments.

\begin{table}[!t]
\caption{Experimentation Results reported by considering various hidden memory units in MDLSTM on Arabic data samples.} \label{tab:exp}
\centering
\begin{tabular}{ |c|c|c|c|c|c| }
\hline

\textbf{MDLSTM Memory block} & \textbf{20} & \textbf{40} & \textbf{60} & \textbf{80} & \textbf{100} \\
\hline
Accuracy  & 39.68 $\pm$ 0.53 & 48.51 $\pm$ 0.22 & 61.74 $\pm$ 1.47 & 73.42 $\pm$ 0.79 & 82.71 $\pm$ 1.13 \\ 
\hline

\end{tabular}
\end{table}

As mentioned earlier, there are few reported efforts in the research direction of Arabic scene text analysis. 
The comparison of presented approach with other cursive scene text analysis is summarized in Table~\ref{tab:comp}.

\begin{table}[!t]
\caption{Experimentation Results reported by considering various hidden memory units in MDLSTM} \label{tab:comp}
\centering
\begin{tabular}{ |c|c|c|c|c| }
\hline

\textbf{Study} & \textbf{Source} &   \textbf{No. of Images} & \textbf{Script} & \textbf{Accuracy} \\
\hline
Sonia .Y et al~\cite{alif}  &  Video text &  $6,532$ & Arabic & $55.03$\% \\ \hline
Cong .Y et al~\cite{md}  &  Camera captured & $6,532$ & Chinese & $75.0$\% \\ \hline
Seonhung .L et al~\cite{edge}  &   Camera captured & $~5,000$ & Korean & $88.0$\% \\ \hline
\textbf{Proposed Technique}  &  \textbf{Camera captured} & \textbf{10,915} & \textbf{Arabic} & \textbf{83.0\%} \\

\hline
\end{tabular}
\end{table}

Most of the presented work in scene text recognition was analyzed on Latin script.
As mentioned in Table~\ref{tab:comp}, other than presented solution for Arabic scene text, there is cursive scene text research proposed, like Chinese and Korean.
The proposed solution presented better accuracy in comparison to other cursive scene text analysis.
The presented work obtained good results and reported best accuracy achieved on Arabic scene text recognition.

\section{Conclusion}
The presented paper has explored the spatial properties associated with an image.
The image properties are exploited by considering various representation of scene text image.
Each image has passed through empirically selected kernels that aim to enhance features of provided samples. 
The cursive nature of Arabic script prompted us to employ MDLSTM, as it is the best suited technique for sequence learning tasks~\cite{urdu1,urdu2,cnn-urdu}.
The dataset was evaluated pyramid wise. 
The presented technique is relatively a novel idea that has not been used before to address the complexities associated with Arabic scripts. 
The proposed work indicated best accuracy in comparison to available Arabic scene text recognition results.
The evaluation of any technique depends on the dataset in question.
One of the possible extension to presented work is to extract statistical features and merge them with presented spatial features for a purpose to perform experiment and evaluate how much merging the said features may impact on the recognition process.

Another important extension of proposed work is to present scale invariant~\cite{sift1} feature extraction technique with the combination of linear pyramids. 
The invariant technique is applied on each pyramid and extract those points which are in common.
This could be an interesting experiment on provided dataset.

\section*{Acknowledgement}
The authors would like to thank Ministry of Education Malaysia and Universiti Teknologi Malaysia for funding this research project.
%
%
%
%

\bibliographystyle{splncs03}

\bibliography{samplepaper}

\end{document}